\documentclass{article}
\usepackage{times}
\usepackage{graphicx} % more modern
\usepackage{subcaption} 
\usepackage{rotating}
\usepackage{natbib}

\usepackage{algorithm}
\usepackage{algorithmic}
\usepackage{amssymb}
\usepackage{booktabs}
\usepackage{amsmath}
\usepackage{etoolbox}
\usepackage{multirow}
\usepackage{xspace}
\usepackage{tikz}
\usetikzlibrary{decorations.pathreplacing}
\usepackage{hyperref}

\usepackage[accepted]{icml2017}
\newcommand{\name}{SWATS\xspace}  

\icmltitlerunning{Improving Generalization Performance by Switching from Adam to SGD}

\begin{document} 

\twocolumn[
\icmltitle{Improving Generalization Performance by Switching from Adam to SGD}
\icmlsetsymbol{equal}{*}

\begin{icmlauthorlist}
\icmlauthor{Nitish Shirish Keskar}{to}
\icmlauthor{Richard Socher}{to}
\end{icmlauthorlist}

\icmlaffiliation{to}{Salesforce Research, Palo Alto, CA -- 94301}
\icmlcorrespondingauthor{Nitish Shirish Keskar}{nkeskar@salesforce.com}
%\icmlcorrespondingauthor{Eee Pppp}{ep@eden.co.uk}

\icmlkeywords{boring formatting information, machine learning, ICML}

\vskip 0.3in
]
%TODO
% Average of N runs



\printAffiliationsAndNotice{}  % leave blank if no need to mention equal contribution

\begin{abstract} 
%Stochastic gradient descent (SGD) and its adaptive variants such as Adam, Adagrad and RMSprop have emerged to be the algorithms of choice for many deep learning training tasks. Despite superior training outcomes in both empirical convergence rate and final objective function value, adaptive methods have been found to generalize poorly compared to SGD. Interestingly, these methods tend to perform well in the initial portion of training but are outperformed by SGD as the algorithms progress. 
Despite superior training outcomes, adaptive optimization methods such as Adam, Adagrad or RMSprop have been found to generalize poorly compared to Stochastic gradient descent (SGD). These methods tend to perform well in the initial portion of training but are outperformed by SGD at later stages of training. 
We investigate a hybrid strategy that begins training with an adaptive method and switches to SGD when appropriate. Concretely, we propose \name, a simple strategy which \textbf{Sw}itches from \textbf{A}dam \textbf{t}o \textbf{S}GD when a triggering condition is satisfied. The condition we propose relates to the projection of Adam steps on the gradient subspace. By design, the monitoring process for this condition adds very little overhead and does not increase the number of hyperparameters in the optimizer. We report experiments on several standard benchmarks such as: ResNet, SENet, DenseNet and PyramidNet for the CIFAR-10 and CIFAR-100 data sets, ResNet on the tiny-ImageNet data set and language modeling with recurrent networks on the PTB and WT2 data sets.
The results show that our strategy is capable of closing the generalization gap between SGD and Adam on a majority of the tasks.
\end{abstract} 

% \section{Introduction}
% \begin{itemize}
% \item SGD is simple but awesome. Lots of theoretical results. 
% \item Adam-family is awesome too. Especially Adam, Don't need to tune hyperparameters, widely useful, rapid initial progress. 
% \item However, Adam doesn't generalize. Attributed to the fact that there is adaptivity.
% \item In this paper,
% \end{itemize}

\section{Introduction}

Stochastic gradient descent (SGD) \cite{robbins1951stochastic} has emerged as one of the most used training algorithms for deep neural networks. Despite its simplicity, SGD performs well empirically across a variety of applications but also has strong theoretical foundations. This includes, but is not limited to, guarantees of saddle point avoidance \cite{lee2016gradient}, improved generalization \cite{hardt2015train,2017arXiv170508292W} and interpretations as Bayesian inference \cite{2017arXiv170404289M}. 

Training neural networks is equivalent to solving the following non-convex optimization problem, 
\begin{align}
\min_{w \in \mathbb{R}^n} & \quad f(w) \nonumber,
\end{align}
where $f$ is a loss function. The iterations of SGD can be described as:
\begin{align}
w_{k} &= w_{k-1} - \alpha_{k-1} \hat{ \nabla} f(w_{k-1}) \nonumber,
\end{align}
where $w_k$ denotes the $k^{th}$ iterate, $\alpha_k$ is a (tuned) step size sequence, also called the learning rate, and $\hat{\nabla} f(w_k)$ denotes the stochastic gradient computed at $w_k$. A variant of SGD (SGDM), that uses the inertia of the iterates to accelerate the training process, has also found to be successful in practice \cite{sutskever2013importance}. The iterations of SGDM can be described as:
\begin{align}
v_{k} &= \beta v_{k-1} + \hat{ \nabla} f(w_{k-1}) \nonumber \\
w_{k} &= w_{k-1} - \alpha_{k-1} v_k \nonumber,
\end{align}
where $\beta \in [0,1)$ is a momentum parameter and $v_0$ is initialized to $0$. 

One disadvantage of SGD is that it scales the gradient uniformly in all directions; this can be particularly detrimental for ill-scaled problems. This also makes the process of tuning the learning rate $\alpha$ circumstantially laborious.

To correct for these shortcomings, several \textit{adaptive} methods have been proposed which diagonally scale the gradient via estimates of the function's curvature. Examples of such methods include Adam \cite{kingma2014adam}, Adagrad \cite{duchi2011adaptive} and RMSprop \cite{tieleman2012lecture}. These methods can be interpreted as methods that use a vector of learning rates, one for each parameter, that are adapted as the training algorithm progresses. This is in contrast to SGD and SGDM which use a scalar learning rate uniformly for all parameters. 

Adagrad takes steps of the form 
\begin{align}
w_{k} &= w_{k-1} - \alpha_{k-1}  \frac{\hat{ \nabla} f(w_{k-1})}{\sqrt{v_{k-1}} + \epsilon}, \quad \text{where} \label{eqn:genadagrad} \\
v_{k-1} &= \sum_{j=1}^{k-1} \hat{\nabla} f(w_j)^2. \nonumber
\end{align}
RMSProp uses the same update rule as \eqref{eqn:genadagrad}, but instead of accumulating $v_k$ in a monotonically increasing fashion, uses an RMS-based approximation instead, i.e.,
\begin{align}
v_{k-1} &= \beta v_{k-2} + (1 - \beta) \hat{\nabla} f(w_{k-1})^2. \nonumber
\end{align}
In both Adagrad and RMSProp, the accumulator $v$ is initialized to $0$. Owing to the fact that $v_k$ is monotonically increasing in each dimension for Adagrad, the scaling factor for $\hat{\nabla} f(w_{k-1})$ monotonically decreases leading to slow progress. RMSProp corrects for this behavior by employing an \textit{average} scale instead of a \textit{cumulative} scale. However, because $v$ is initialized to $0$, the initial updates tend to be noisy given that the scaling estimate is biased by its initialization.  This behavior is rectified in Adam by employing a bias correction. Further, it uses an exponential moving average for the step in lieu of the gradient. Mathematically, the Adam update equation can be represented as:
\begin{align}
w_{k} &= w_{k-1} - \alpha_{k-1} \cdot \frac{\sqrt{1 - \beta_2^k}}{1-\beta_1^k} \cdot  \frac{m_{k-1}}{\sqrt{v_{k-1}} + \epsilon}, \quad \text{where} \label{eqn:adameps} \\
m_{k-1} &= \beta_1 m_{k-2} + (1 - \beta_1) \hat{\nabla} f(w_{k-1}), \nonumber \\
v_{k-1} &= \beta_2 v_{k-2} + (1 - \beta_2) \hat{\nabla} f(w_{k-1})^2. \label{eqn:adambeta2}
\end{align}

Adam has been used in many applications owing to its competitive performance and its ability to work well despite minimal tuning \cite{karpathyadam}. Recent work, however, highlights the possible inability of adaptive methods to perform on par with SGD when measured by their ability to generalize \cite{2017arXiv170508292W}.

Furthermore, the authors also show that for even simple quadratic problems, adaptive methods find solutions that can be orders-of-magnitude worse at generalization than those found by SGD(M).

Indeed, for several state-of-the-art results in language modeling and computer vision, the optimizer of choice is SGD \cite{merityRegOpt,loshchilov2016sgdr,he2015deep}. Interestingly however, in these and other instances, Adam outperforms SGD in both training and generalization metrics in the initial portion of the training, but then the performance stagnates. This motivates the investigation of a strategy that combines the benefits of Adam, viz. good performance with default hyperparameters and fast initial progress, and the generalization properties of SGD. Given the insights of  \citet{2017arXiv170508292W} which suggest that the lack of generalization performance of adaptive methods stems from the non-uniform scaling of the gradient, a natural \textit{hybrid} strategy would begin the training process with Adam and switch to SGD when appropriate. To investigate this further, we propose \name, a simple strategy that combines the best of both worlds by \textbf{Sw}itching from \textbf{A}dam \textbf{t}o \textbf{S}GD. The switch is designed to be automatic and one that does not introduce any more hyper-parameters. The choice of not adding additional hyperparameters is deliberate since it allows for a fair comparison between Adam and \name. Our experiments on several architectures and data sets suggest that such a strategy is indeed effective. 

% \section{Related Work}
% \begin{itemize}
% \item Lots of work on Adam recently
% \item NC-Adam
% \item AMSGrad
% \item \url{https://arxiv.org/pdf/1711.05101.pdf}
% \item these propose new variants or regularization strategies but that doesn't really fix the problem of Wilson given that the adaptivity is what hurts. 
% \item \url{https://www.preferred-networks.jp/docs/imagenet_in_15min.pdf}
% \item GNMT
% \item Contributions:
% \end{itemize}
Several attempts have been made at improving the convergence and generalization performance of Adam. The closest to our proposed approach is \cite{zhang2017normalized} in which the authors propose ND-Adam, a variant of Adam which preserves the gradient direction by a nested optimization procedure. This, however, introduces an additional hyperparameter along with the $(\alpha,\beta_1,\beta_2)$ used in Adam. Further, empirically, this adaptation sacrifices the rapid initial progress typically observed for Adam.  In \citet{anonymous2018on}, the authors investigate Adam and ascribe the poor generalization performance to training issues arising from the non-monotonic nature of the steps. The authors propose a variant of Adam called AMSGrad which monotonically reduces the step sizes and possesses theoretical convergence guarantees. Despite these guarantees, we empirically found the generalization performance of AMSGrad to be similar to that of Adam on problems where a generalization gap exists between Adam and SGD. We note that in the context of the hypothesis of \citet{2017arXiv170508292W}, all of the aforementioned methods would still yield poor generalization given that the scaling of the gradient is non-uniform. 

The idea of switching from an adaptive method to SGD is not novel and has been explored previously in the context of machine translation \cite{wu2016google} and  ImageNet training \cite{akiba2017extremely}. \citet{wu2016google} use such a mixed strategy for training and tune both the switchover point and the learning rate for SGD after the switch. \citet{akiba2017extremely} use a similar strategy but use a convex combination of RMSProp and SGD steps whose contributions and learning rates are tuned.

In our strategy, the switchover point and the SGD learning rate are both learned as a part of the training process. We monitor a projection of the Adam step on the gradient subspace and use its exponential average as an estimate for the SGD learning rate after the switchover. Further, the switchover is triggered when no change in this monitored quantity is detected. We describe this strategy in detail in Section~\ref{sec:name}. In Section \ref{sec:exp}, we describe our experiments comparing Adam, SGD and \name on a host of benchmark problems. Finally, in Section \ref{sec:discon}, we present ideas for future research and concluding remarks. We conclude this section by emphasizing the goal of this work is less to propose a new training algorithm but rather to empirically investigate the viability of hybrid training for improving generalization.

%TODO: LR not step size

\section{\name}
\label{sec:name}
% \begin{itemize}
% \item Begin by making an observation about Adam. Clipping experiment. 
% \item motivation for accumulating
% \item tikz figure for how the steps are projected
% \item actual algorithm
% \item side comment about convergence
% \end{itemize}

To investigate the generalization gap between Adam and SGD, let us consider the training of the CIFAR-10 data set \cite{krizhevsky2009learning} on the DenseNet architecture \cite{iandola2014densenet}. This is an example of an instance where a significant generalization gap exists between Adam and SGD. We plot the performance of Adam and SGD on this task but also consider a variant of Adam which we call Adam-Clip$(p,q)$. Given $(p,q)$ such that $p<q$, the iterates for this variant take on the form
\begin{align*}
w_{k} &= w_{k-1} - \\ &\text{\text{clip}}\left(\frac{\sqrt{1 - \beta_2^k}}{1-\beta_1^k}  \frac{\alpha_{k-1}}{\sqrt{v_{k-1}}+\epsilon} , p\cdot \alpha_{sgd}, q\cdot \alpha_{sgd}\right) m_{k-1}.
\end{align*}

\begin{figure}
\begin{center}
%\framebox[4.0in]{$\;$}
\includegraphics[width=\columnwidth]{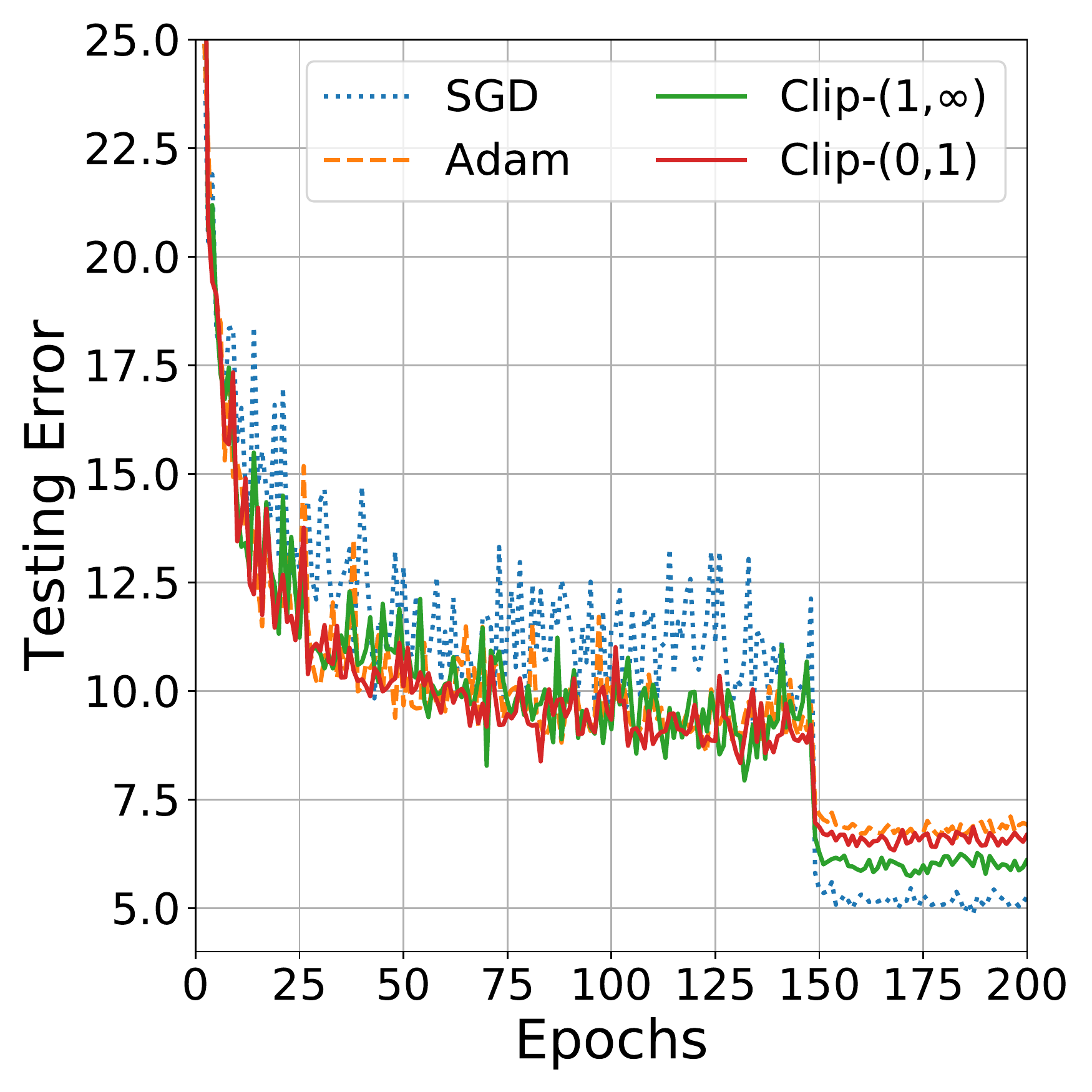}
\end{center}
\caption{ Training the DenseNet architecture on the CIFAR-10 data set with four optimizers: SGD, Adam, Adam-Clip$(1,\infty)$ and Adam-Clip$(0, 1)$. SGD achieves the best testing accuracy while training with Adam leads to a generalization gap of roughly $2\%$. Setting a minimum learning rate for each parameter of Adam partially closes the generalization gap. \label{fig:densenet-clip}}
\end{figure}

Here, $\alpha_{sgd}$ is the tuned value of the learning rate for SGD that leads to the best performance for the same task. The function $\text{{clip}}(x,a,b)$ clips the vector $x$ element-wise such that the output is constrained to be in $[a,b]$. Note that Adam-Clip$(1,1)$ would correspond to SGD. The network is trained using Adam, SGD and two variants: Adam-Clip$(1,\infty)$, Adam-Clip$(0, 1)$ with tuned learning rates for 200 epochs, reducing the learning rate by $10$ after $150$ epochs.  The goal of this experiment is to investigate the effect of constraining the large and small step sizes that Adam implicitly learns, i.e., $\frac{\sqrt{1 - \beta_2^k}}{1-\beta_1^k} \frac{\alpha_{k-1}}{\sqrt{v_{k-1}}+\epsilon}$, on the generalization performance of the network. We present the results in Figure~\ref{fig:densenet-clip}.

As seen from Figure~\ref{fig:densenet-clip}, SGD converges to the expected testing error of $\approx 5\%$ while Adam stagnates in performance at around $\approx 7\%$ error. We note that fine-tuning of the learning rate schedule (primarily the initial value, reduction amount and the timing) did not lead to better performance. Also, note that the rapid initial progress of Adam relative to SGD. This experiment is in agreement with the experimental observations of \citet{2017arXiv170508292W}. Interestingly, Adam-Clip$(0, 1)$ has no tangible effect on the final generalization performance while Adam-Clip$(1, \infty)$ partially closes the generalization gap by achieving a final accuracy of $\approx 6\%$. We observe similar results for several architectures, data sets and modalities whenever a generalization gap exists between SGD and Adam. This stands as evidence that the step sizes learned by Adam could circumstantially be too small for effective convergence. 
This observation regarding the need to lower-bound the step sizes of Adam is similar to the one made in \citet{anonymous2018on}, where the authors devise a one-dimensional example in which infrequent but large gradients are not emphasized sufficiently causing the non-convergence of Adam. 

\begin{figure}[t]
\begin{center}
%\framebox[4.0in]{$\;$}
\includegraphics[width=\columnwidth]{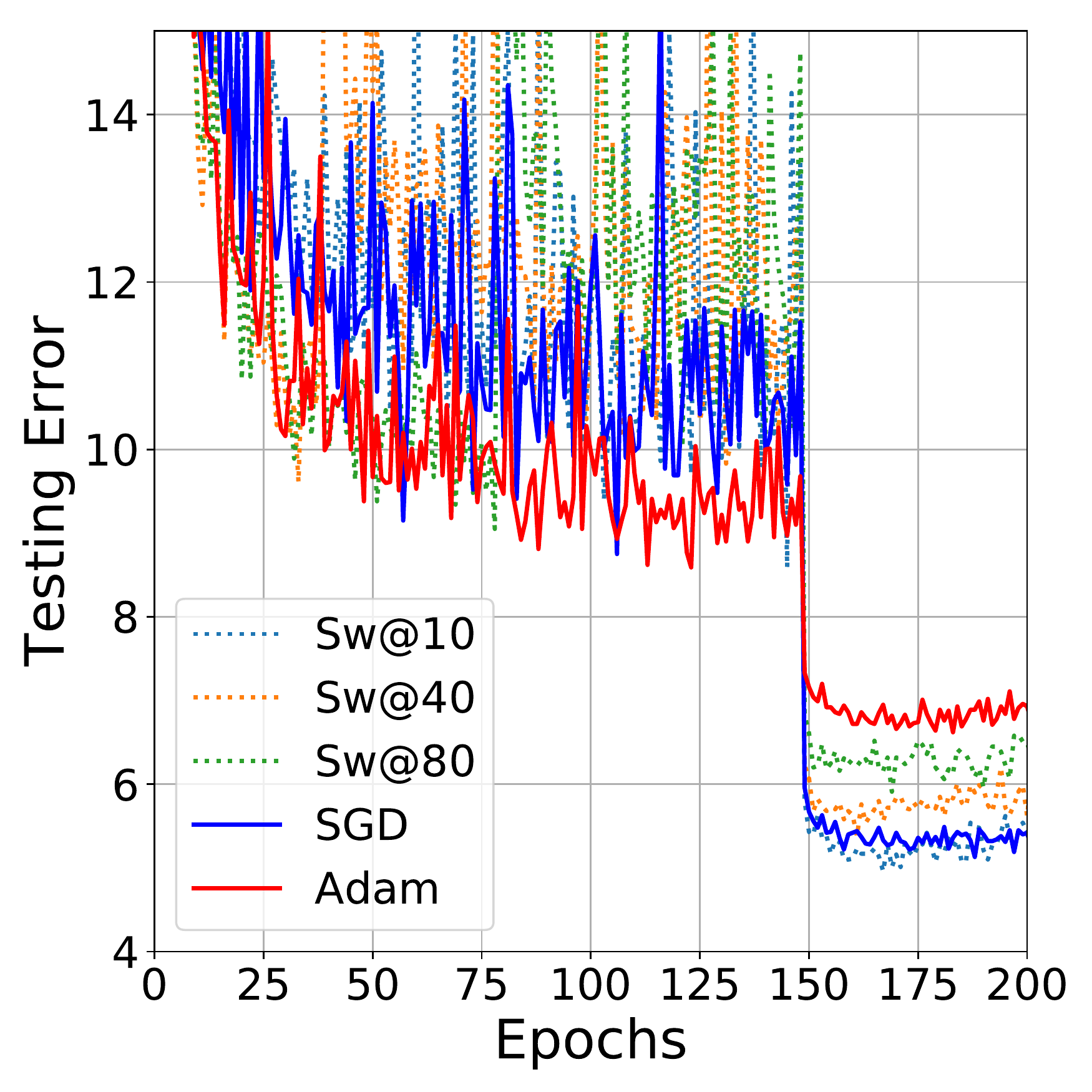}
\end{center}
\caption{ Training the DenseNet architecture on the CIFAR-10 data set using Adam and switching to SGD with learning rate with learning rate $0.1$ and momentum $0.9$ after $(10,40,80)$ epochs; the switchover point is denoted by Sw@ in the figure. Switching early enables the model to achieve testing accuracy comparable to SGD but switching too late in the training process leads to a generalization gap similar to Adam. \label{fig:densenet-switches}}
\end{figure}

Given the potential insufficiency of Adam, even when constraining one side of the accumulator, we consider \textit{switching} to SGD once we have reaped the benefits of Adam's rapid initial progress. This raises two questions: (a) when to switch over from Adam to SGD, and (b) what learning rate to use for SGD after the switch. Assuming that the learning rate of SGD after the switchover is tuned, we found that switching too late does not yield generalization improvements while switching too early may cause the hybrid optimizer to not benefit from Adam's initial progress. Indeed, as shown in Figure~\ref{fig:densenet-switches}, switching after $10$ epochs leads to a learning curve very similar to that of SGD, while switching after $80$ epochs leads to inferior testing accuracy of $\approx 6.5\%$. To investigate the efficacy of a hybrid strategy whilst ensuring no increase in the number of hyperparameters (a necessity for fair comparison with Adam), we propose \name, a strategy that automates the process of switching over by determining both the switchover point and the learning rate of SGD after the switch. 

\subsection{Learning rate for SGD after the switch}
Consider an iterate $w_k$ with a stochastic gradient $g_k$ and a step computed by Adam, $p_k$. For the sake of simplicity, assume that $p_k\neq 0$ and $p_k^T g_k <0$. This is a common requirement imposed on directions to derive convergence \cite{nocedal2006numerical}. In the case when $\beta_1=0$ for Adam, i.e., no first-order exponential averaging is used, this is trivially true since $$p_k = - \underbrace{\frac{\sqrt{1 - \beta_2^{k+1}}}{1-\beta_1^{k+1}}  \frac{\alpha_{k}}{\sqrt{v_{k}}+\epsilon}}_{:= \text{diag}(H_k) } g_k,t$$ with $H_k \succ 0$ where $\text{diag}(A)$ denotes the vector constructed from the diagonal of $A$. Ordinarily, to train using Adam, we would update the iterate as:
$$w_{k+1} = w_k + p_k.$$
To determine a feasible learning rate for SGD, $\gamma_k$, we propose solving the subproblem for finding $\gamma_k$
%$$\gamma_k = \min_{\gamma \in \mathbb{R}^+} \| (w_k + p_k ) - (w_k - \gamma g_k) \|_2^2.$$
$$\text{proj}_{-\gamma_k g_k} p_k = p_k$$
where $\text{proj}_a b$ denotes the orthogonal projection of $a$ onto $b$. 
This scalar optimization problem can be solved in closed form to yield:
$$ \gamma_k = \frac{p_k^T p_k}{-p_k^T g_k},$$ 
since
$$p_k=\text{proj}_{-\gamma_k g_k} p_k = -\gamma_k \frac{g_k^T p_k}{p_k^T p_k} p_k$$
implies the above equality. 

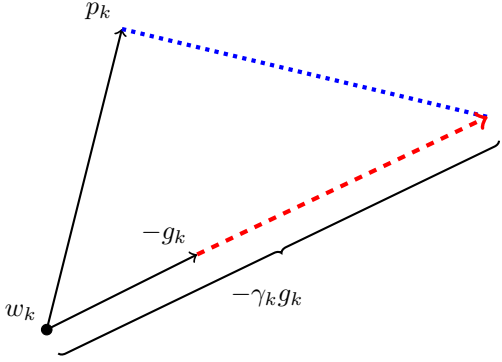
\begin{figure}
\begin{tikzpicture}
%\draw (-2,0) -- (2,0);
\filldraw[black] (0,0) circle (2pt);
\draw[->,thick] (0,0) node[above left] {$w_k$} -- (1,4) node[above left] {$p_k$} ;
\draw[->,thick] (0,0) -- (2,1) node[above left] {$-g_k$} ;
%\draw[->,dashed,blue,thick] (0,0) -- (-2,-1) node[above left] {$g_k$} ;
\draw[ultra thick,dotted, blue] (1,4) -- (5.86,2.83);
\draw[dashed,red,ultra thick,->] (2,1) -- (5.86,2.83);
\draw[decoration={brace,mirror,raise=10pt},decorate,thick]
  (0,0) -- (5.86,2.83) node[midway,below=20pt] {$-\gamma_k g_k$};
\end{tikzpicture}
\caption{Illustrating the learning rate for SGD ($\gamma_k$) estimated by our proposed projection given an iterate $w_k$, a stochastic gradient $g_k$ and the Adam step $p_k$. \label{fig:tikzproj}}
\end{figure}

Geometrically, this can be interpreted as the scaling necessary for the gradient that leads to its projection on the Adam step $p_k$ to be $p_k$ itself; see Figure~\ref{fig:tikzproj}. Note that this is not the same as an orthogonal projection of $p_k$ on $-g_k$. Empirically, we found that an orthogonal projection consistently underestimates the SGD learning rate necessary, leading to much smaller SGD steps. Indeed, the $\ell_2$ norm of an orthogonally projected step will always be lesser than or equal to that of $p_k$, which is undesirable given our needs. The non-orthogonal projection proposed above does not suffer from this problem, and empirically we found that it estimates the SGD learning rate well. A simple scaling rule of $\gamma_k = \frac{\|p\|}{\|g\|}$ was also not found to be successful. We attribute this to the fact that a scaling rule of this form ignores the relative importance of the coordinate directions and tends to amplify the importance of directions with a large step $p$ but small first-order importance $g$, and vice versa. 

Note again that if no momentum ($\beta_1=0$) is employed in Adam, then necessarily $\gamma_k > 0$ since $H_k \succ 0$. We should mention in passing that in this case $\gamma_k$ is equivalent to the reciprocal of the Rayleigh Quotient of $H_k^{-1}$ with respect to the vector $p_k$. 

Since $\gamma_k$ is a noisy estimate of the scaling needed, we maintain an exponential average initialized at $0$, denoted by $\lambda_k$ such that 
$$\lambda_k = \beta_2 \lambda_{k-1}  + (1 - \beta_2) \gamma_k.$$
We use $\beta_2$ of Adam, see \eqref{eqn:adambeta2}, as the averaging coefficient since this reuse avoids another hyperparameter and also because the performance is relatively invariant to fine-grained specification of this parameter.

\subsection{Switchover Point}
Having answered the question of what learning rate $\lambda_k$ to choose for SGD after the switch, we now discuss when to switch. We propose checking a simple, yet powerful, criterion:
%$\lambda_k$ answers the question of what learning rate to choose for SGD after the switch. To decide \textit{when} to switch, we propose checking a simple criterion:
\begin{align}
\Big|\frac{\lambda_k}{1 - \beta_2^k} - \gamma_k \Big| < \epsilon,
\end{align}
at every iteration with $k>1$. The condition compares the bias-corrected exponential averaged value and the current value ($\gamma_k$). The bias correction is necessary to prevent the influence of the zero initialization during the initial portion of training. Once this condition is true, we switch over to SGD with learning rate $\Lambda := \frac{\lambda_k}{ (1 - \beta_2^k)}$. We also experimented with more complex criteria including those involving monitoring of gradient norms. However, we found that this simple un-normalized criterion works well across a variety of different applications. 

In the case when $\beta_1>0$, we switch to SGDM with learning rate $(1-\beta_1) \Lambda$ and momentum parameter $\beta_1$. The $(1-\beta_1)$ factor is the common momentum correction. Refer to Algorithm~\ref{alg:name} for a unified view of the algorithm. The text in blue denotes operations that are also present in Adam. 
%pp = 1 + 16 = 17 , pg = 2 + 4 = 6, pp/pg = 17/6

%%%%%%% ALGORITHM %%%%%%%
\begin{algorithm}[t]
\caption{\name}
\label{alg:madam}
{\bf Inputs:} Objective function $f$, initial point $w_0$, learning rate $\alpha=10^{-3}$, accumulator coefficients $(\beta_1,\beta_2) = (0.9, 0.999)$, $\epsilon=10^{-9}$, phase=\texttt{Adam}.

\begin{algorithmic}[1]
\STATE {\color{blue} Initialize $k\gets 0$, $m_k \gets 0$, $a_k \gets 0$,} $\lambda_k \gets 0$
\WHILE{ {\color{blue}stopping criterion not met}}
\STATE {\color{blue} $k = k + 1$}
\STATE {\color{blue} Compute stochastic gradient $g_k=\hat{\nabla} f(w_{k-1})$ }
\IF{phase = \texttt{SGD}}
\STATE $v_{k} = \beta_1 v_{k-1} + g_k $
\STATE $w_{k} = w_{k-1} - (1 - \beta_1) \Lambda v_k $
\STATE \bf continue
\ENDIF
\STATE {\color{blue} $m_k = \beta_1 m_{k-1} + (1 - \beta_1) g_k$}
\STATE {\color{blue} $a_k = \beta_2 a_{k-1} + (1 - \beta_2) g_k^2$}
\STATE {\color{blue} $p_k = -\alpha_k \frac{\sqrt{1 - \beta_2^k}}{1 - \beta_1^k} \frac{m_k}{\sqrt{a_k} + \epsilon} $ }
\STATE {\color{blue} $w_k = w_k + p_k$}
\IF{ $p_k^T g_k \neq 0$}
\STATE  $\gamma_k = \frac{p_k^T p_k}{- p_k^T g_k} $
\STATE $\lambda_k = \beta_2 \lambda_{k-1}  + (1 - \beta_2) \gamma_k$
\IF{$k>1$ and $|\frac{\lambda_k}{(1 - \beta_2^k)} - \gamma_k| < \epsilon$}
\STATE phase = \texttt{SGD}
\STATE $v_k = 0$
\STATE $\Lambda = \lambda_k / (1 - \beta_2^k)$
\ENDIF
\ELSE
\STATE $\lambda_k = \lambda_{k-1}$
\ENDIF
\ENDWHILE
\end{algorithmic}
{\bf return }  $w_k$
\label{alg:name}
\end{algorithm}
%%%%%%% END OF ALGORITHM

%TODO: We briefly comment on the convergence guarantees of such an algorithm. 

\section{Numerical Results}
To demonstrate the efficacy of our approach, we present numerical experiments comparing the proposed strategy with Adam and SGD. We consider the problems of image classification and language modeling. 

For the former, we experiment with four architectures: ResNet-32 \cite{he2015deep}, DenseNet \cite{iandola2014densenet}, PyramidNet \cite{DPRN}, and SENet \cite{hu2017squeeze} on the CIFAR-10 and CIFAR-100 data sets \cite{krizhevsky2009learning}. The goal is to classify images into one of 10 classes for CIFAR-10 and 100 classes for CIFAR-100. The data sets contain $50000$ $32\times 32$ RGB images in the training set and $10000$ images in the testing set. We choose these architectures given their superior performance on several image classification benchmarking tasks. For a large-scale image classification experiment, we experiment with the Tiny-ImageNet data set\footnote{\url{https://tiny-imagenet.herokuapp.com/}} on the ResNet-18 architecture \cite{he2015deep}. This data set is a subset of the ILSVRC 2012 data set \cite{imagenet_cvpr09} and contains $200$ classes with $500$ $224\times 224$ RGB images per class in the training set and $50$ per class in the validation and testing sets. We choose this data set given that it is a good proxy for the performance on the larger ImageNet data set. 

We also present results for word-level language modeling where the task is to take as inputs a sequence of words and predict the next word. We choose this task because of its broad importance, the  inherent difficulties that arise due to long term dependencies \cite{hochreiter1997long}, and since it is a proxy for other sequence learning tasks such as machine translation \cite{bahdanau2014neural}. We use the Penn Treebank (PTB) \cite{mikolov2011rnnlm} and the larger WikiText-2 (WT-2) \cite{merity2016pointer} data sets and experimented with the AWD-LSTM and AWD-QRNN architectures. In the case of SGD, we clip the gradients to a norm of $0.25$ while we perform no such clipping for Adam and \name. We found that the performance of SGD deteriorates without clipping and that of Adam and \name with. The AWD-LSTM architecture uses a multi-layered LSTM network with learned embeddings while the AWD-QRNN replaces the expensive LSTM layer by the cheaper QRNN layer \cite{Bradbury2016} which uses convolutions instead of recurrences. The model is regularized with DropConnect \cite{dropconnect} on the hidden-to-hidden connections as well as other strategies such as weight decay, embedding-softmax weight tying, activity regularization and temporal activity regularization. We refer the reader to \cite{merity2016pointer} for additional details regarding the data sets including the sizes of the training, validation and testing sets, size of the vocabulary, and source of the data.  

For our experiments, we tuned the learning rate of all optimizers, and report the best-performing configuration in terms of generalization. The learning rate of Adam and \name were chosen from a grid of $\{0.0005,0.0007,$ $0.001,0.002,0.003,0.004,0.005\}.$ For both optimizers, we use the (default) recommended values $(\beta_1, \beta_2)=(0.9,0.999)$. Note that this implies that, in all cases, we switch from Adam to SGDM with a momentum coefficient of $0.9$. For tuning the learning rate for the SGD(M) optimizer, we first coarsely tune the learning rate on a logarithmic scale from $10^{-3}$ to $10^2$ and then fine-tune the learning rate. For all cases, we experiment with and without employing momentum but don't tune this parameter ($\beta=0.9$). We found this overall procedure to perform better than a generic grid-search or hyperparameter optimization given the vastly different scales of learning rates needed for different modalities. For instance, SGD with learning rate $0.7$ performed best for the DenseNet task on CIFAR-10 but for the PTB language modeling task using the LSTM architecture, a learning rate of $50$ for SGD was necessary. Hyperparameters such as batch size, dropout probability, $\ell_2$-norm decay etc. were chosen to match the recommendations of the respective base architectures. We trained all networks for a total of $300$ epochs and reduced the learning rate by $10$ on epochs $150$, $225$ and $262$. This scheme was surprisingly powerful at obtaining good performance across the different modalities and architectures. The experiments were coded in PyTorch\footnote{\url{pytorch.org}} and conducted using job scheduling on $16$ NVIDIA Tesla K80 GPUs for roughly 3 weeks.

\label{sec:exp}
% \begin{itemize}
% \item lots of tables and pictures.
% \item If Adam is insufficient, improves upon it. If Adam is good enough, matches it. 
% \item learning rate obtained is similar to that found by tuning SGD. might be worth restarting?
% \item benefits from no tuning
% \item in almost all cases, default hyperparameters MAdam wins
% \item meta-analsis of which hyperparameter performed best and when the switchover happened and with what value
% \end{itemize}

\begin{figure*}
    \centering
    \begin{subfigure}[b]{0.24\textwidth}
        \includegraphics[width=\columnwidth]{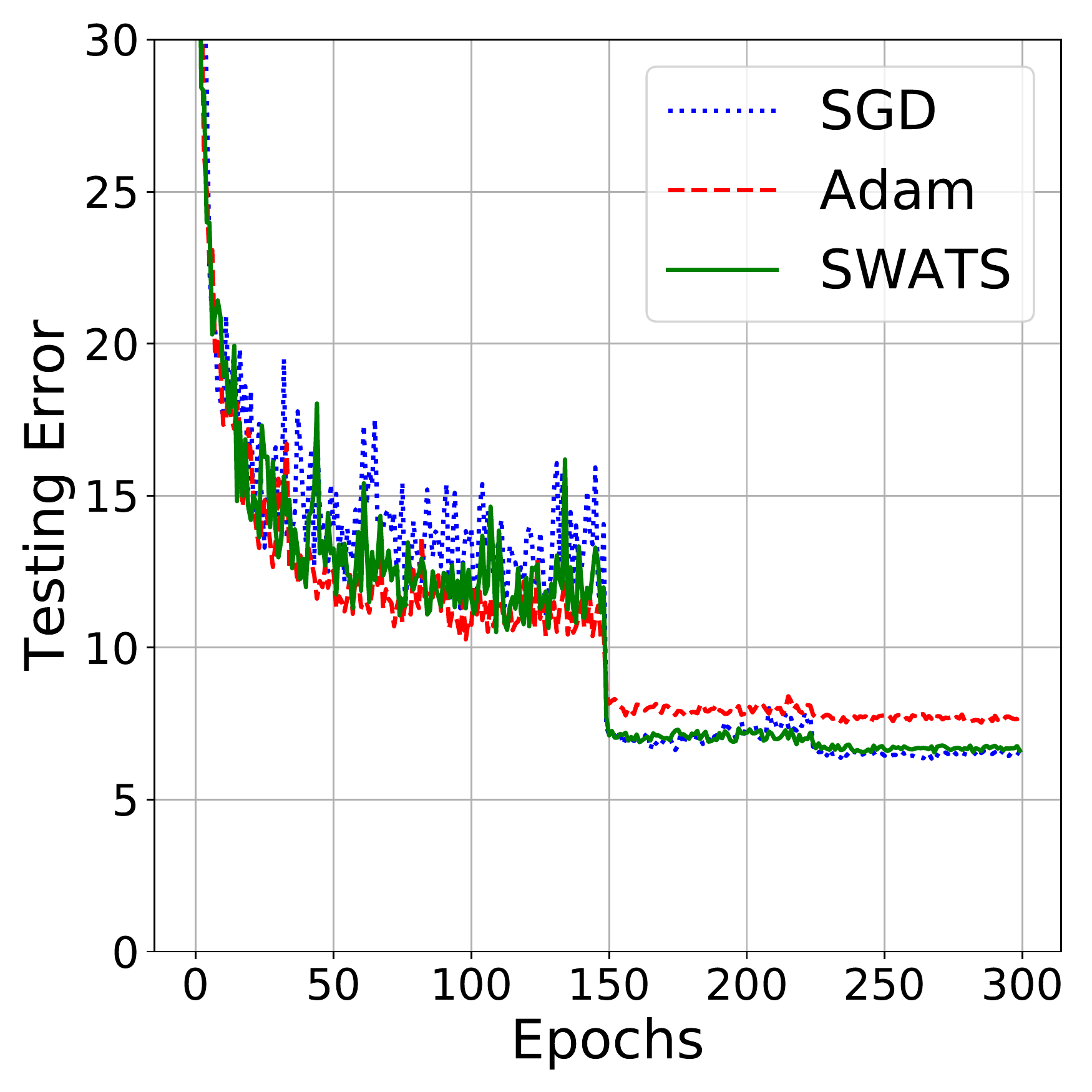}
        \centering
        \caption{ResNet-32 --- CIFAR-10}
        \label{fig:gull}
    \end{subfigure}
    \begin{subfigure}[b]{0.24\textwidth}
        %\fbox{\rule[-.5cm]{0cm}{4cm} \rule[-.5cm]{4cm}{0cm}}
        \includegraphics[width=\columnwidth]{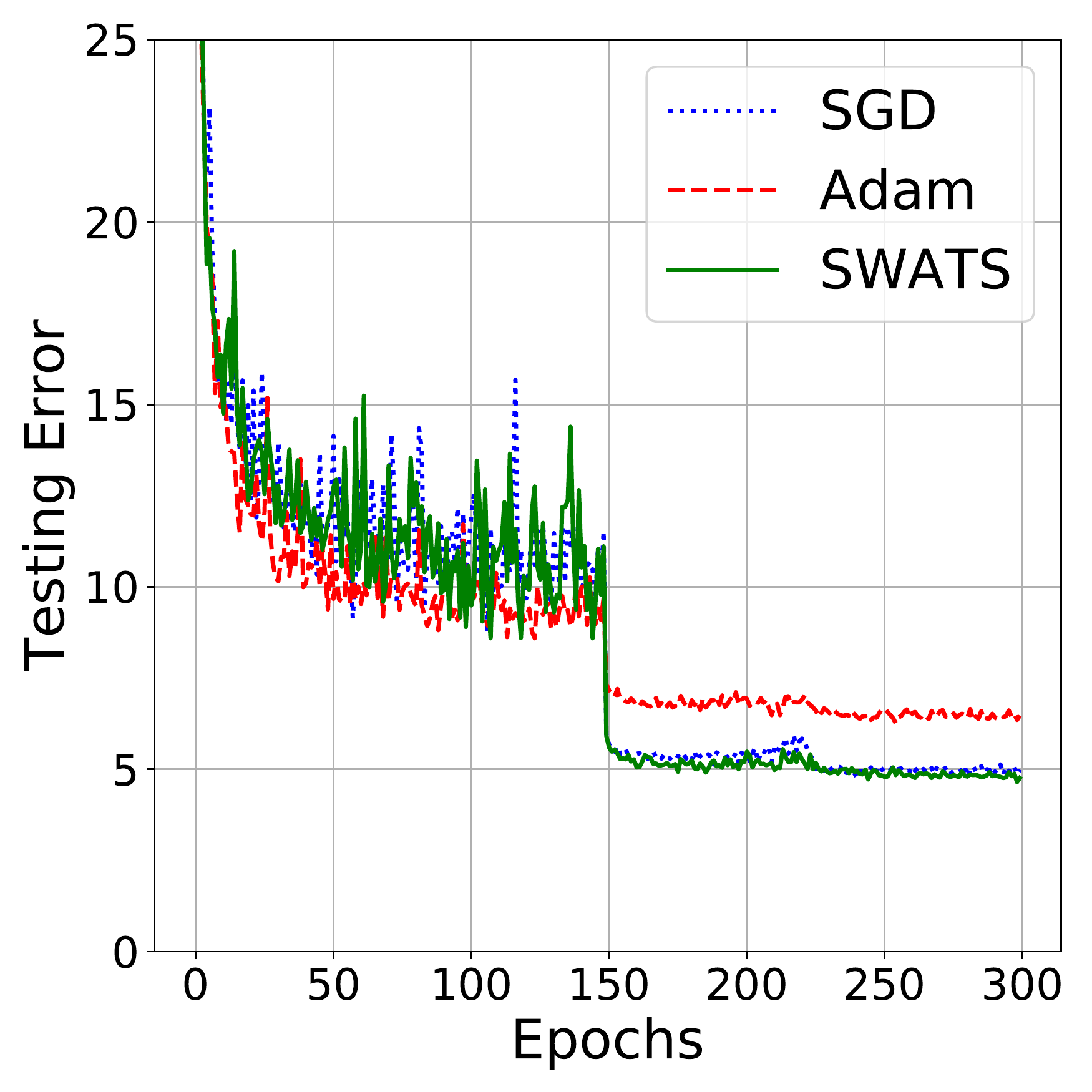}
        \centering
        \caption{DenseNet --- CIFAR-10}        
        \label{fig:tiger}
    \end{subfigure}
    \begin{subfigure}[b]{0.24\textwidth}
        \includegraphics[width=\columnwidth]{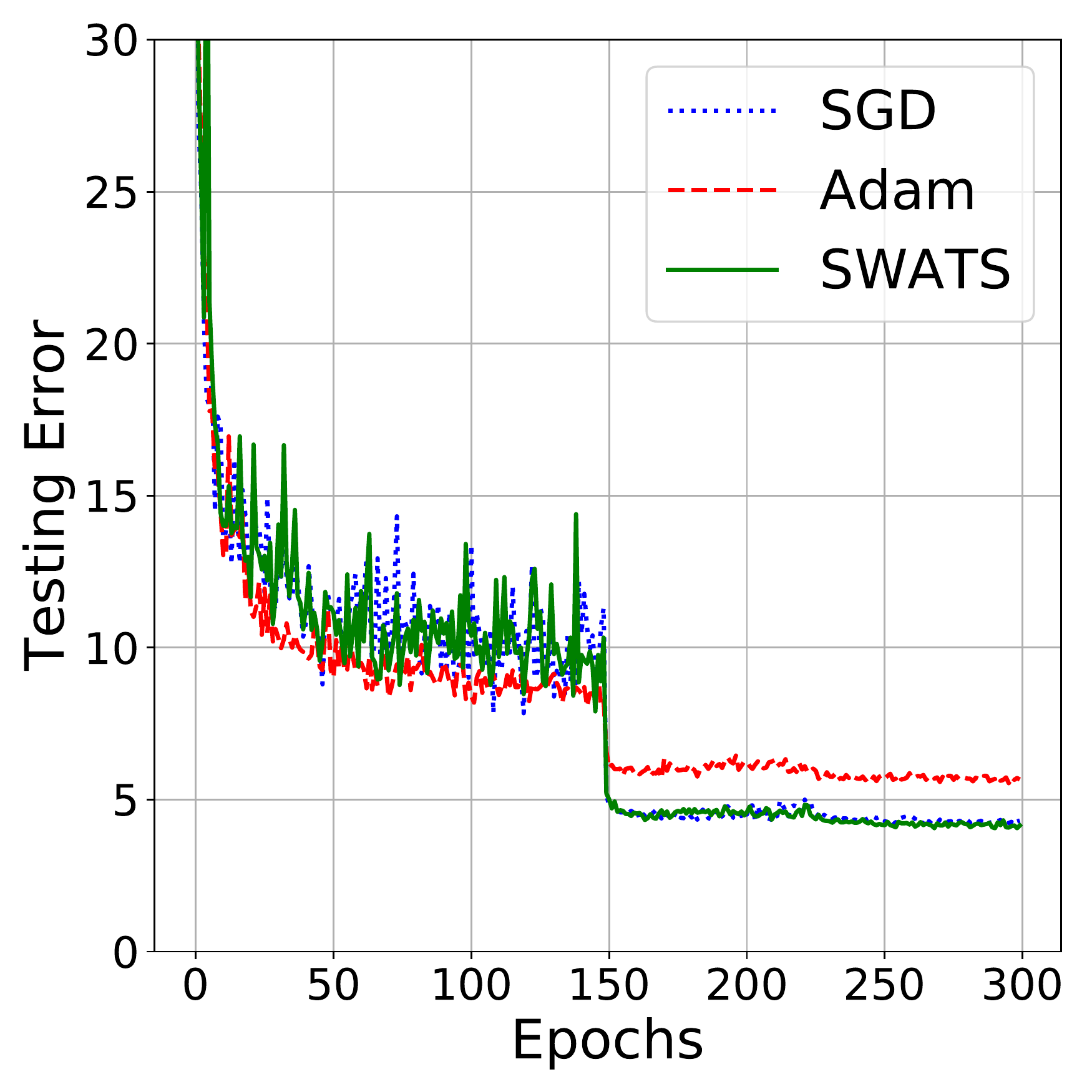}
        \centering
        \caption{PyramidNet --- CIFAR-10}
        \label{fig:gull}
    \end{subfigure}
    \begin{subfigure}[b]{0.24\textwidth}
        \includegraphics[width=\columnwidth]{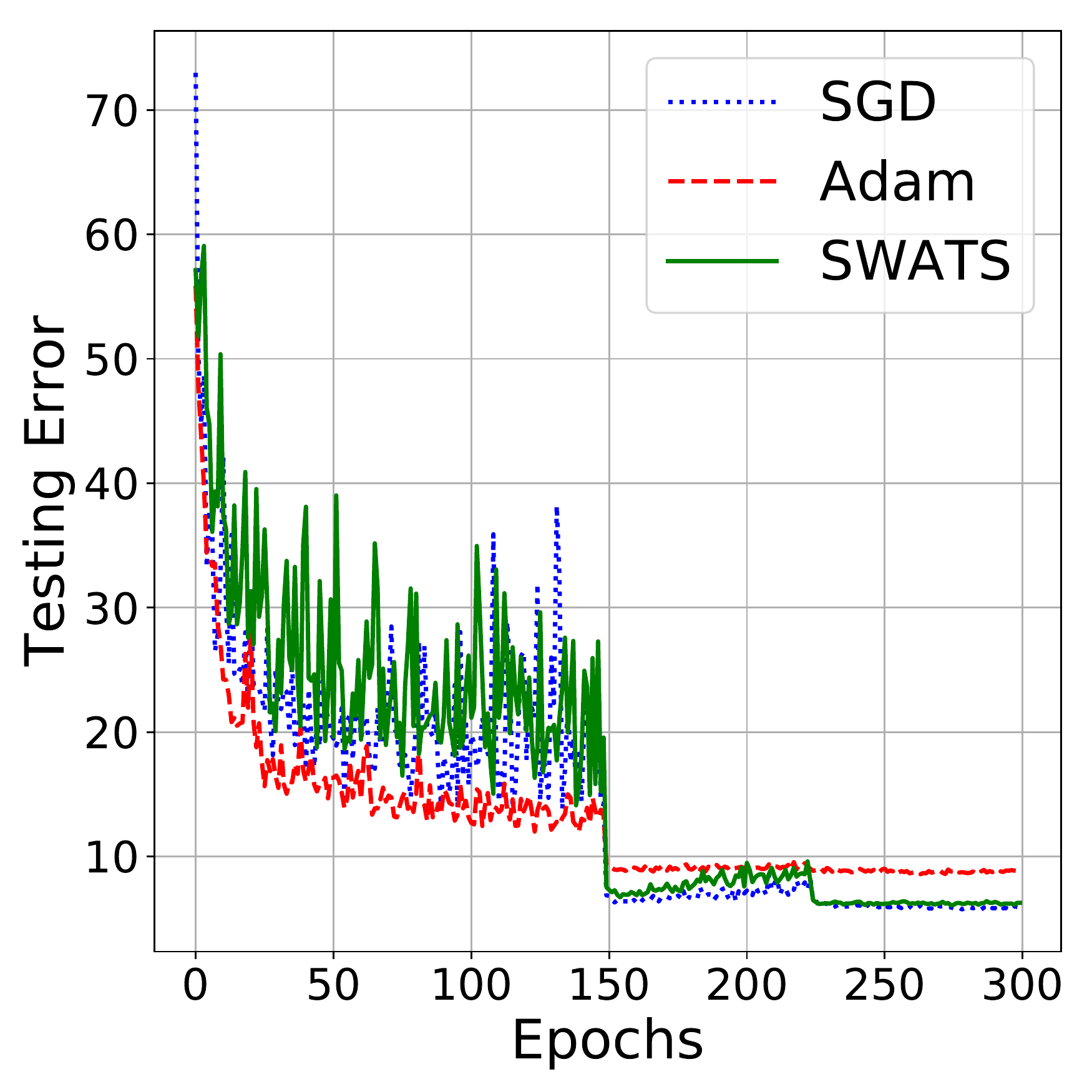}
        \centering
        \caption{SENet --- CIFAR-10}        
        \label{fig:tiger}
    \end{subfigure}
	\\[5ex]
    \begin{subfigure}[b]{0.24\textwidth}
        \includegraphics[width=\columnwidth]{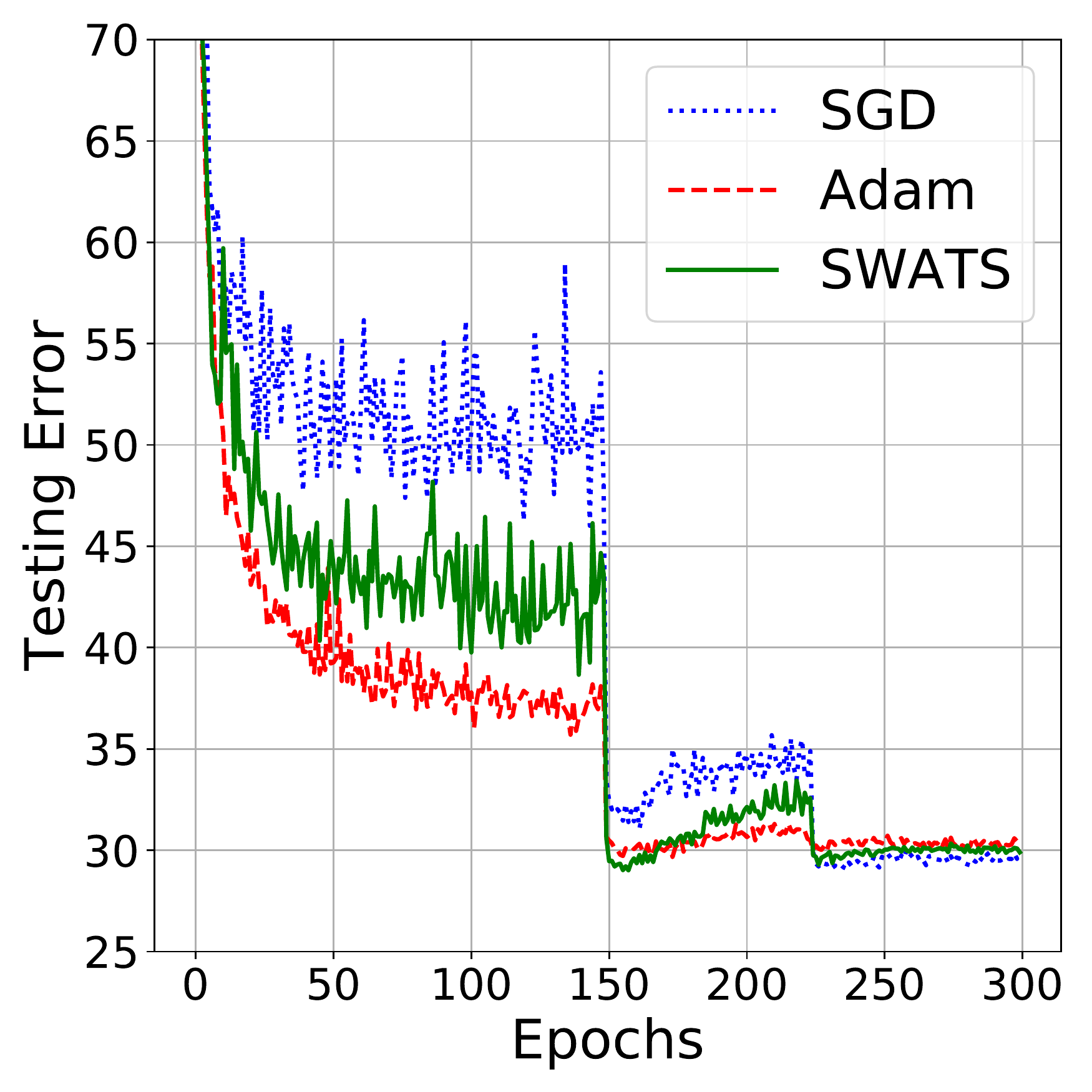}
        \centering
        \caption{ResNet-32 --- CIFAR-100}
        \label{fig:gull}
    \end{subfigure}
    \begin{subfigure}[b]{0.24\textwidth}
        \includegraphics[width=\columnwidth]{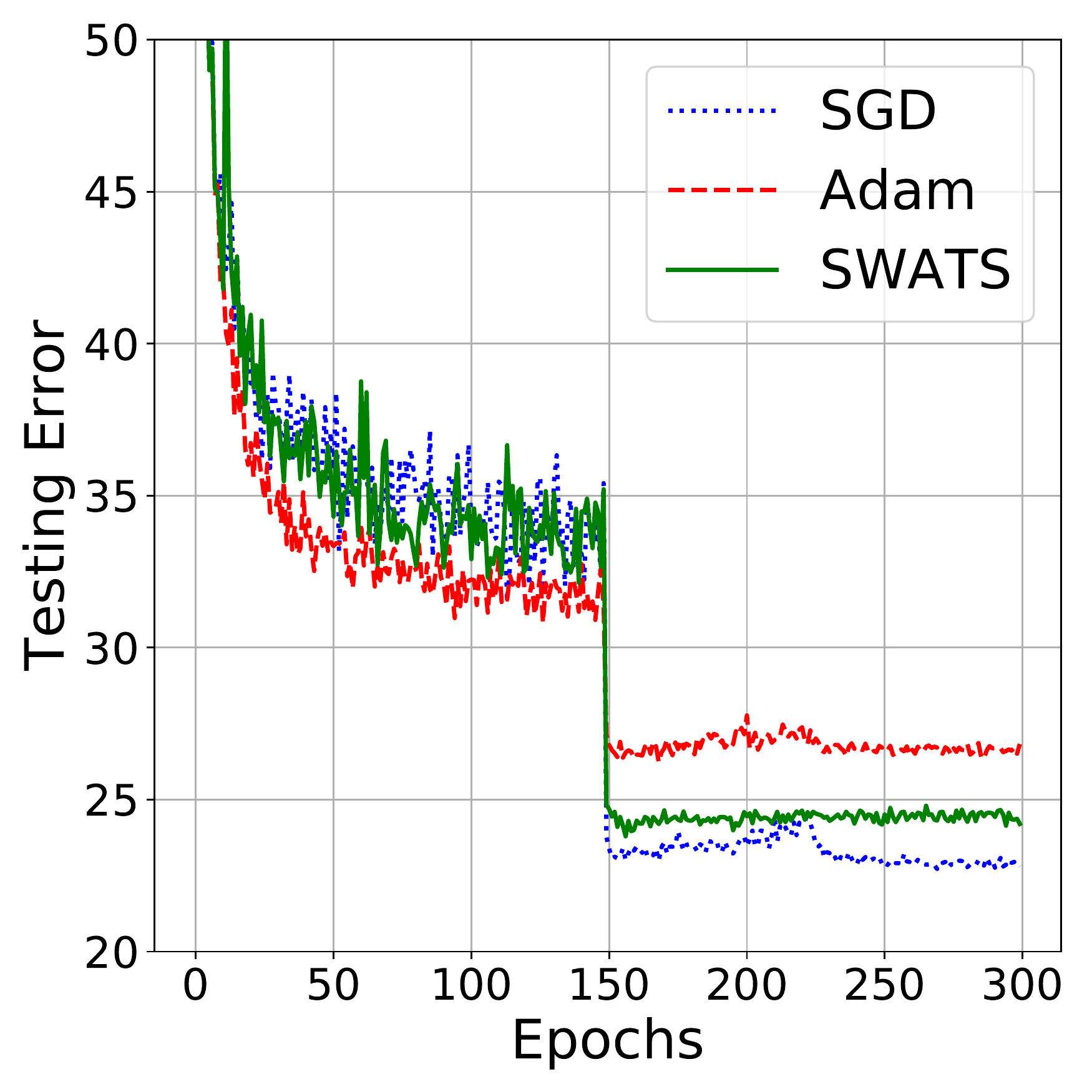}
        \centering
        \caption{DenseNet --- CIFAR-100}        
        \label{fig:tiger}
    \end{subfigure}
    \begin{subfigure}[b]{0.24\textwidth}
        \includegraphics[width=\columnwidth]{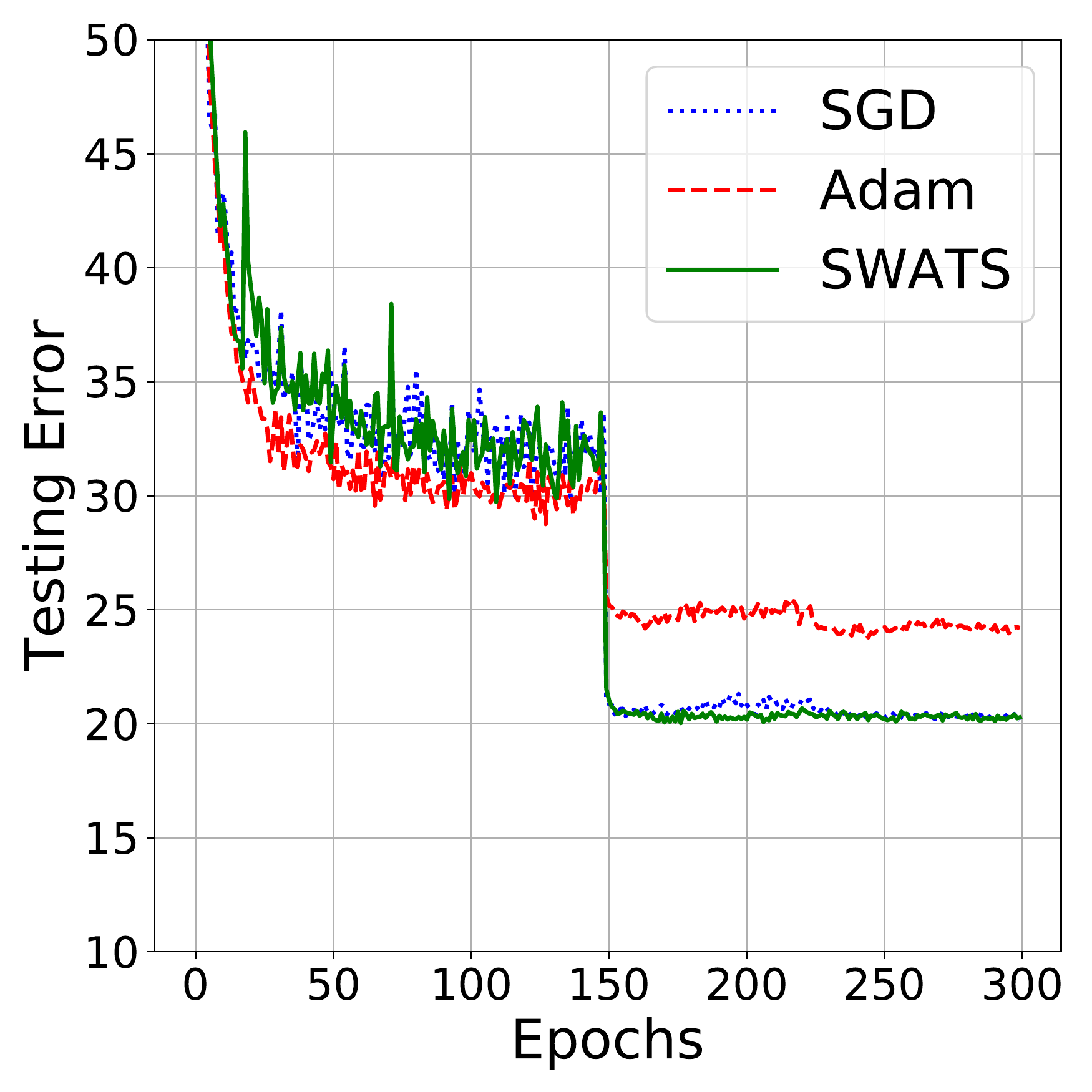}
        \centering
        \caption{PyramidNet --- CIFAR-100}
        \label{fig:gull}
    \end{subfigure}
    \begin{subfigure}[b]{0.24\textwidth}
        \includegraphics[width=\columnwidth]{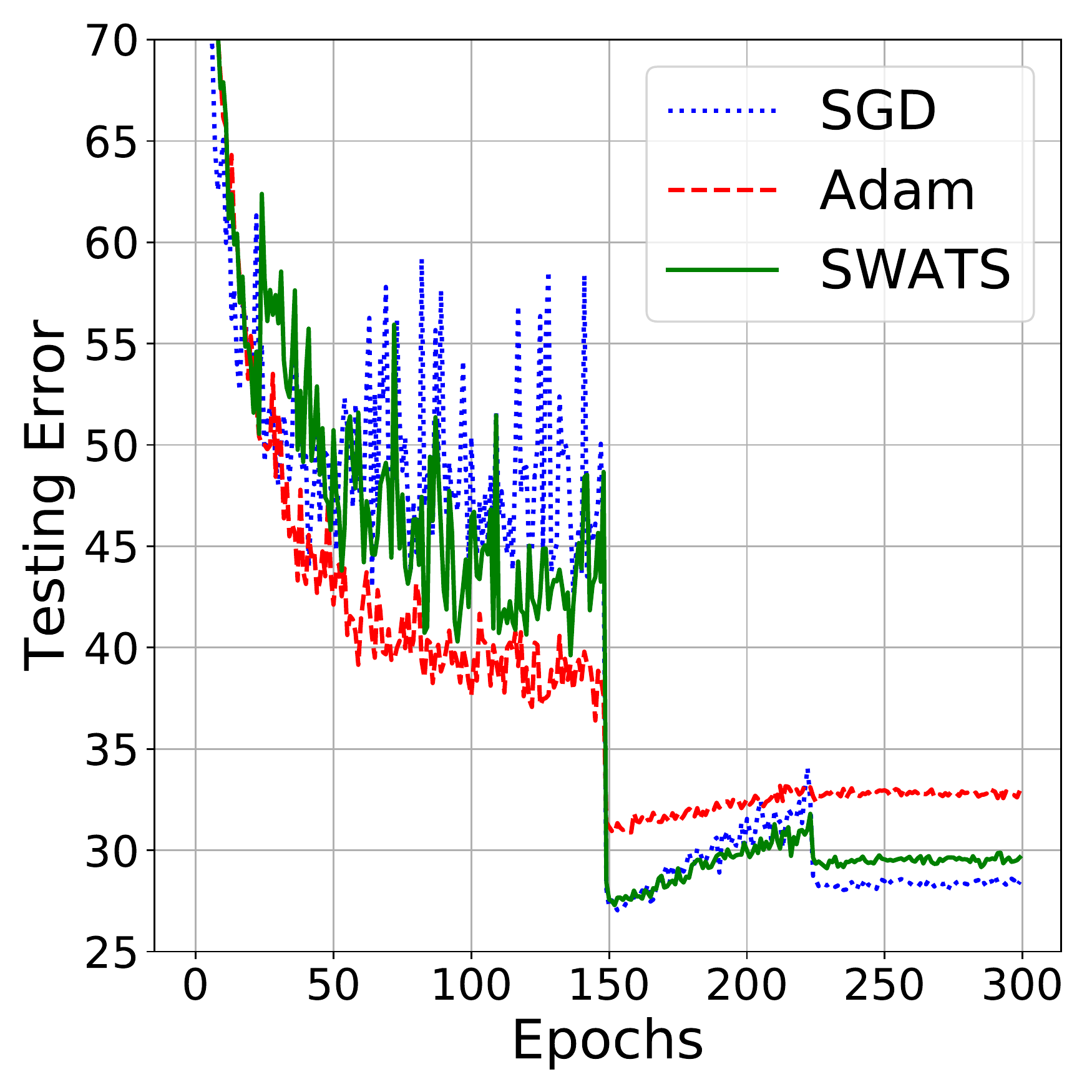}
        \centering
        \caption{SENet --- CIFAR-100}        
        \label{fig:tiger}
    \end{subfigure}
    \caption{Numerical experiments comparing SGD(M), Adam and \name with tuned learning rates on the ResNet-32, DenseNet, PyramidNet and SENet architectures on CIFAR-10 and CIFAR-100 data sets.}\label{fig:cifars}
\end{figure*}

\begin{figure}
\centering
\includegraphics[width=\columnwidth]{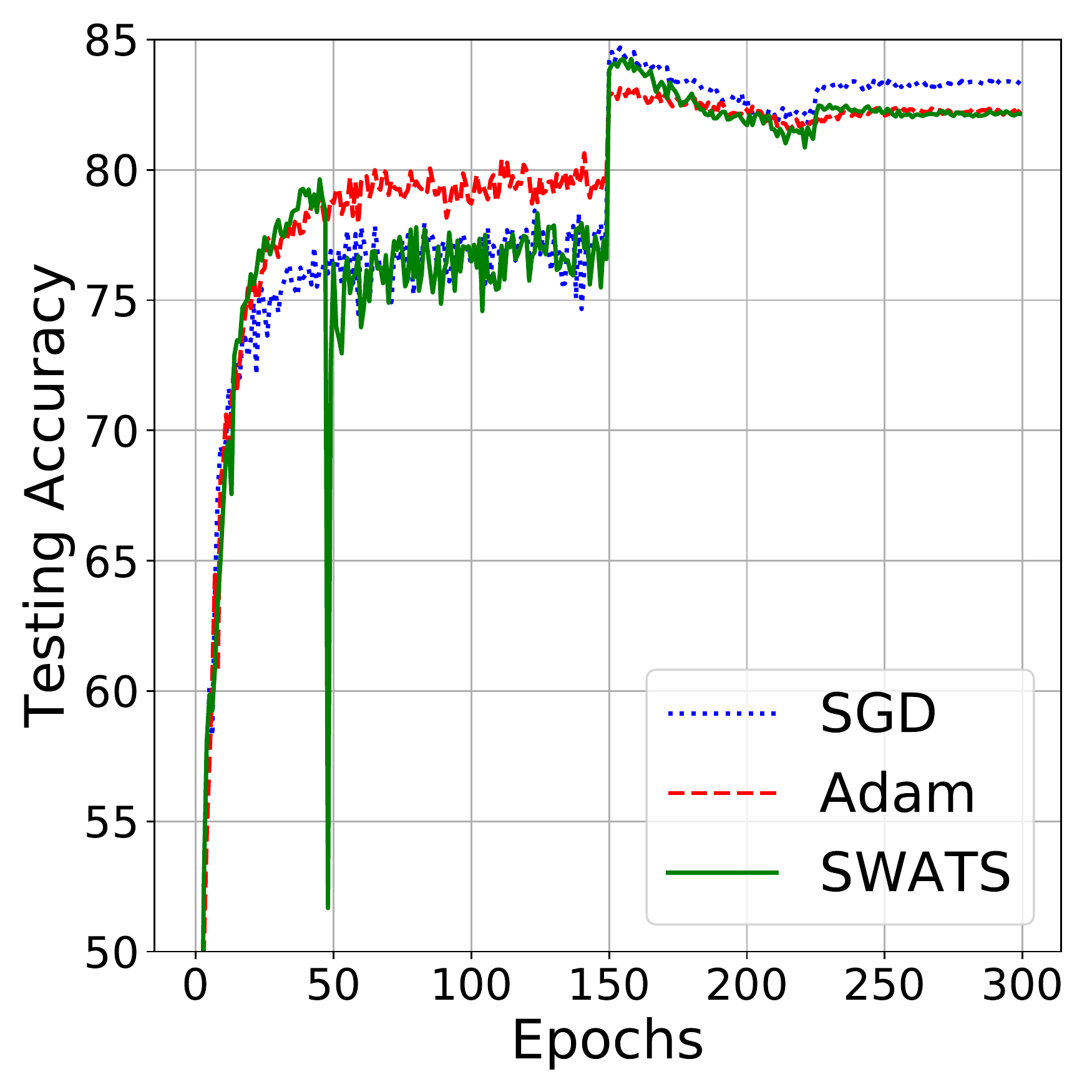}
\caption{Numerical experiments comparing SGD(M), Adam and \name with tuned learning rates on the ResNet-18 architecture on the Tiny-ImageNet data set.}
\label{fig:tiny}
\end{figure}

\begin{figure*}
    \centering
    \begin{subfigure}[b]{0.24\textwidth}
        \includegraphics[width=\columnwidth]{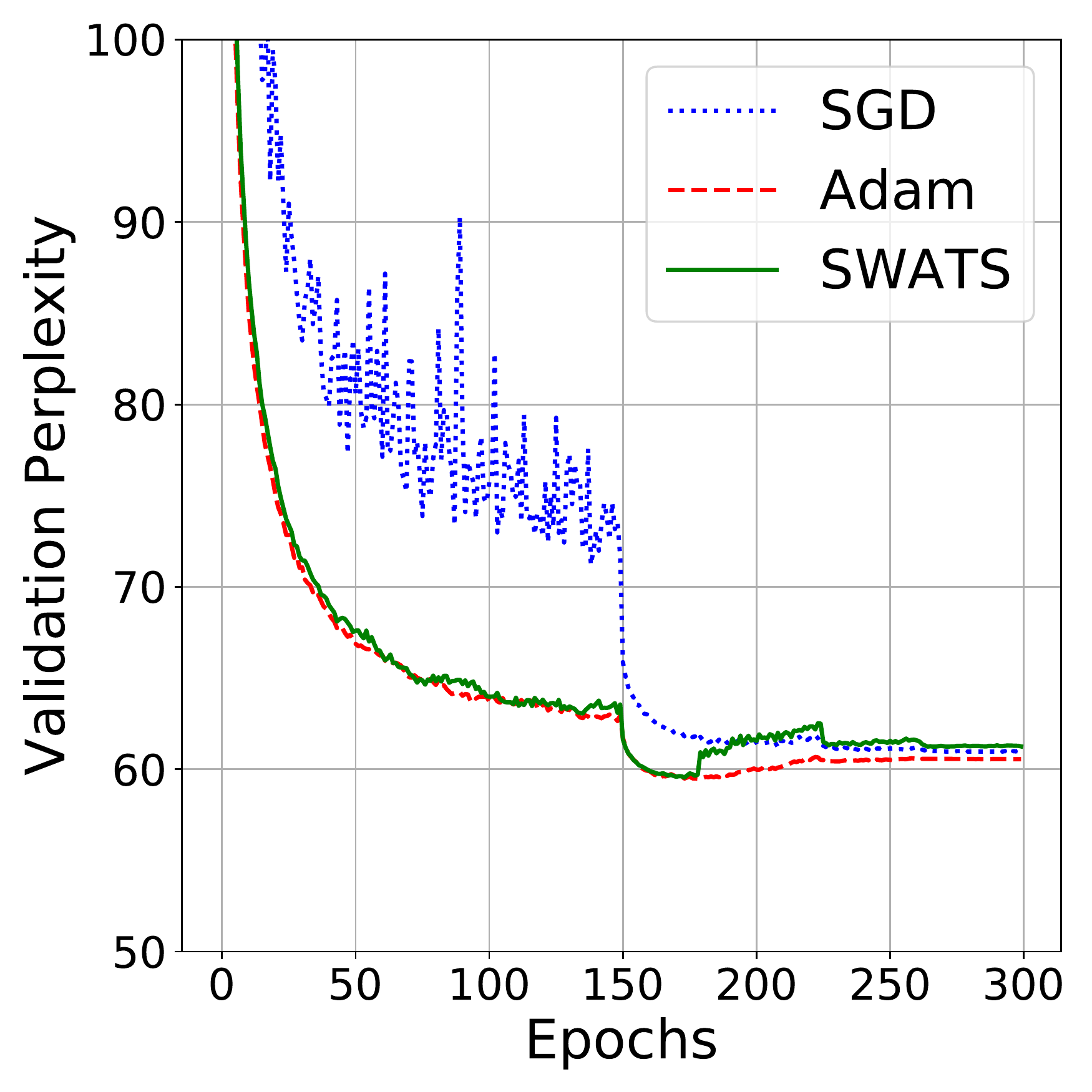}
        \centering
        \caption{LSTM --- PTB}
        \label{fig:gull}
    \end{subfigure}
    \begin{subfigure}[b]{0.24\textwidth}
        \includegraphics[width=\columnwidth]{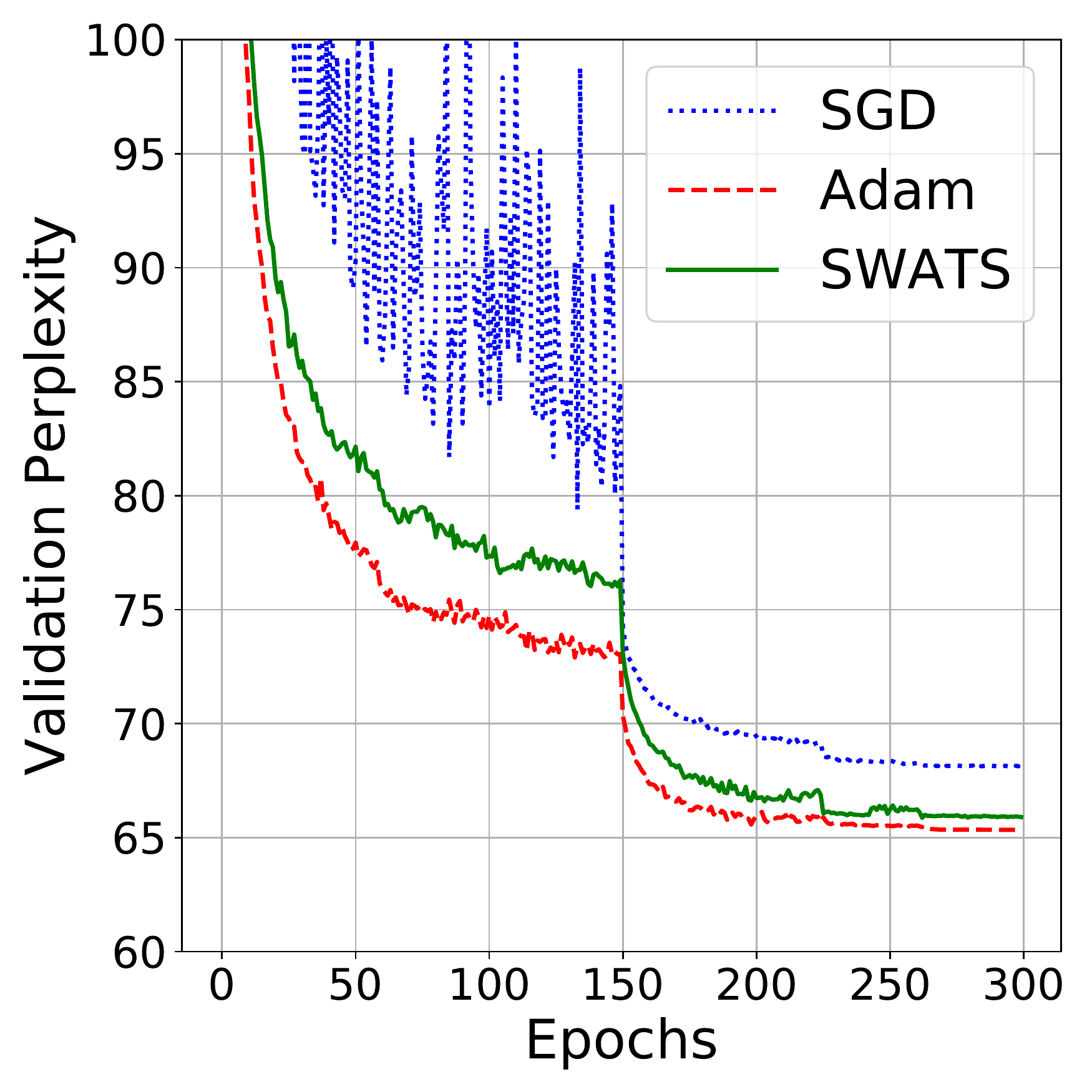}
        \centering
        \caption{LSTM --- WT2}        
        \label{fig:tiger}
    \end{subfigure}
    \begin{subfigure}[b]{0.24\textwidth}
        \includegraphics[width=\columnwidth]{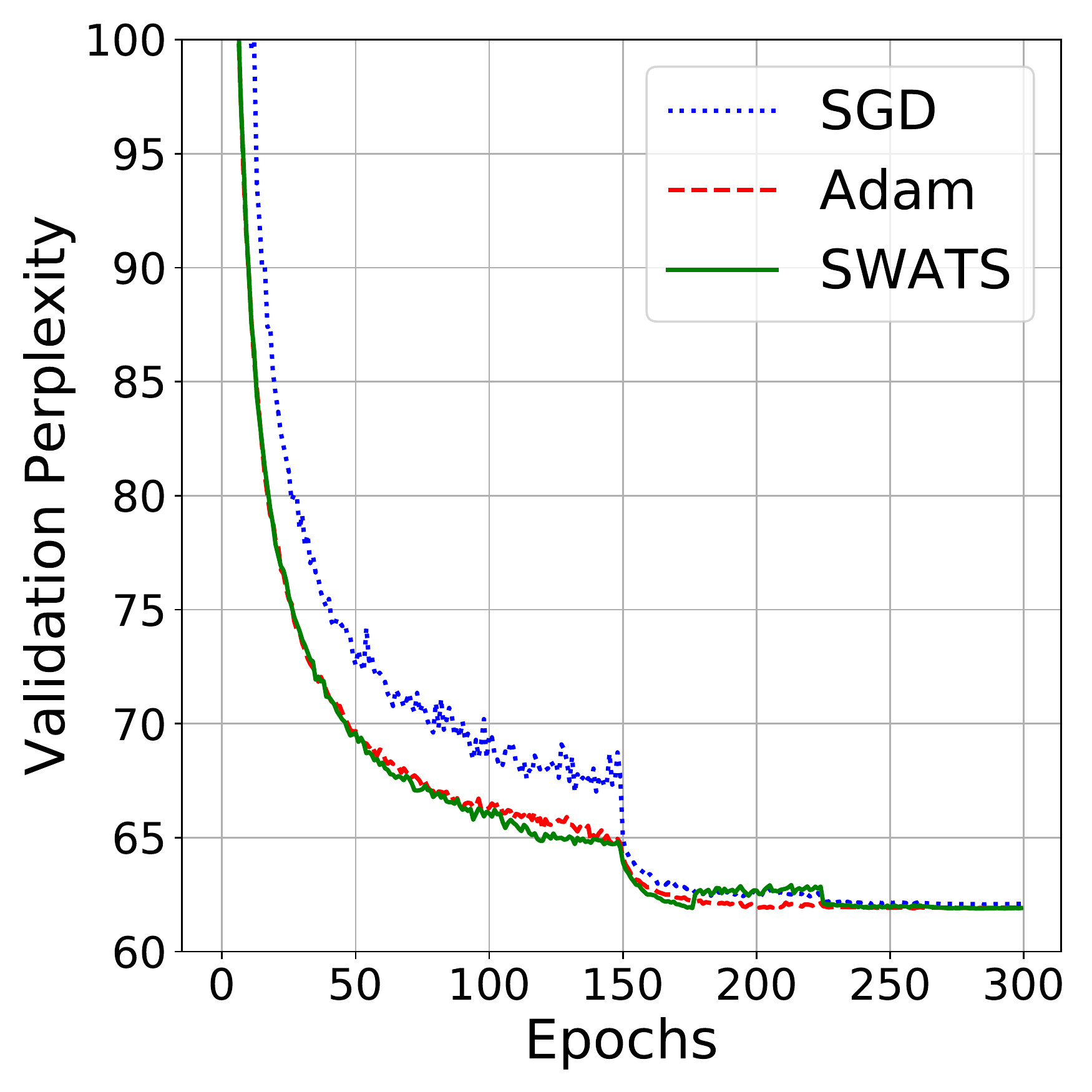}
        \centering
        \caption{QRNN --- PTB}
        \label{fig:gull}
    \end{subfigure}
    \begin{subfigure}[b]{0.24\textwidth}
        \includegraphics[width=\columnwidth]{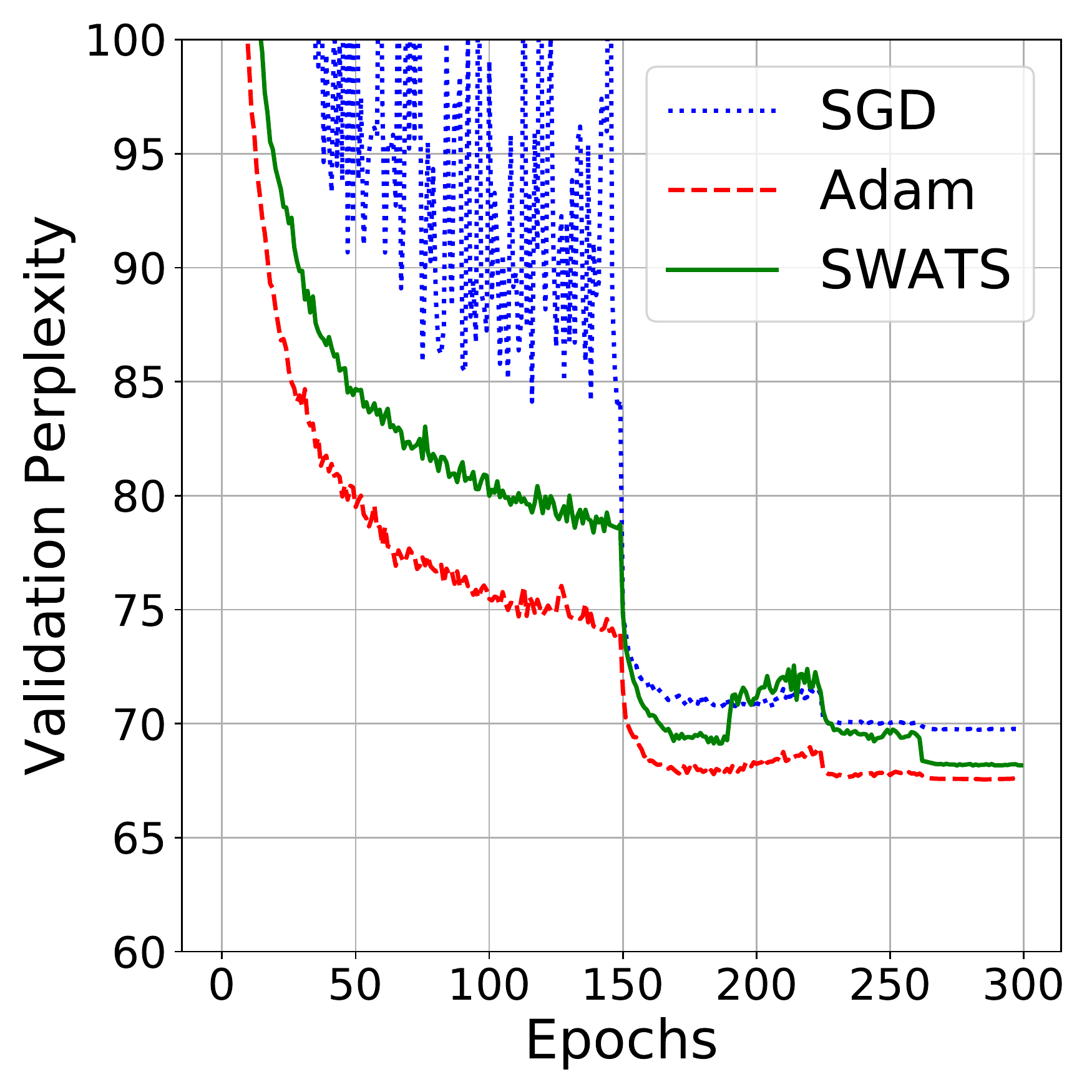}
        \centering
        \caption{QRNN --- WT2}        
        \label{fig:tiger}
    \end{subfigure}
    \caption{Numerical experiments comparing SGD(M), Adam and \name with tuned learning rates on the AWD-LSTM and AWD-QRNN architectures on PTB and WT-2 data sets.\label{fig:awdlstmlm}}
\end{figure*}

\begin{figure}
    \centering
    \begin{subfigure}[b]{0.49\columnwidth}
        \includegraphics[width=\columnwidth]{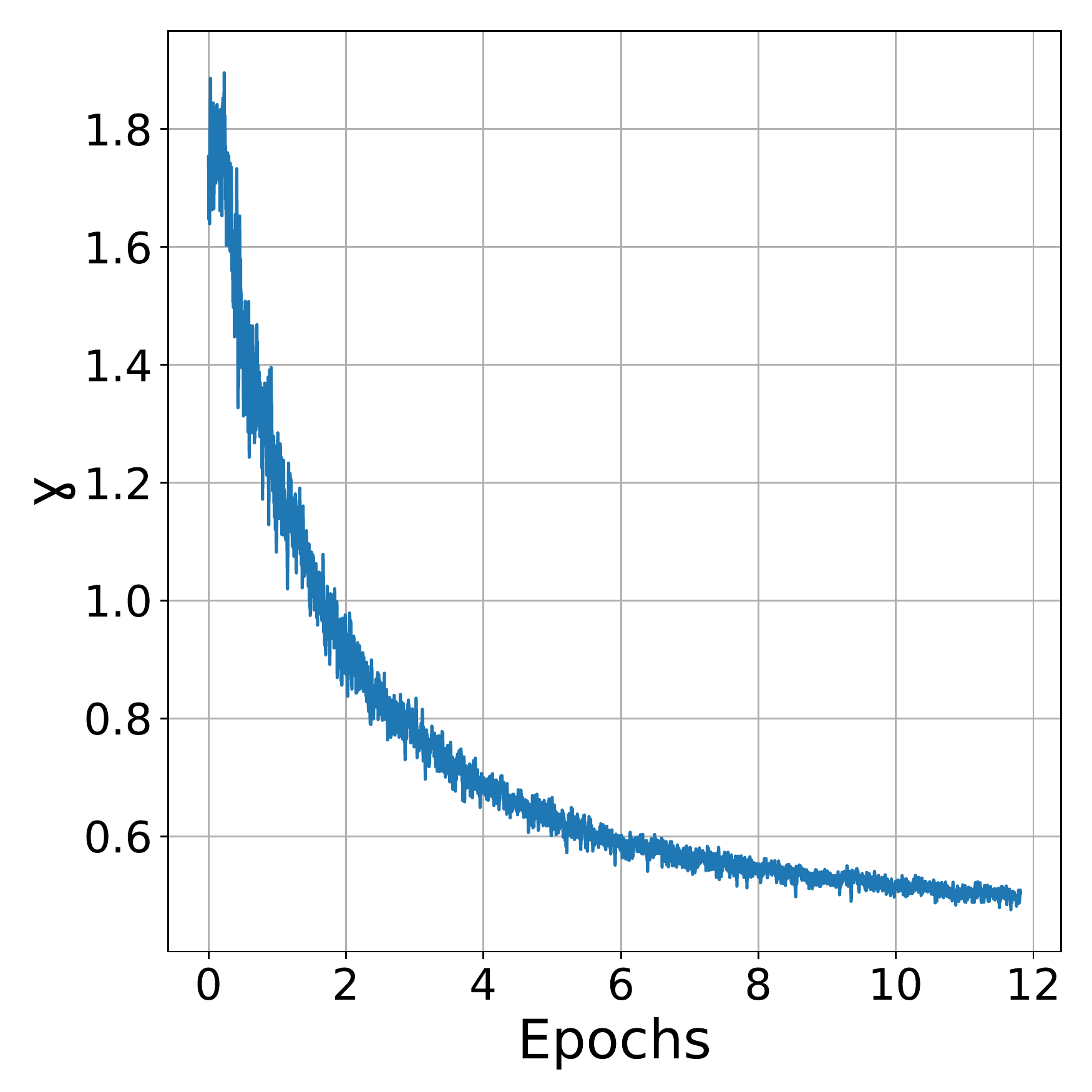}
        \centering
        \caption{DenseNet --- CIFAR-100}
        \label{fig:gull}
    \end{subfigure}
    \begin{subfigure}[b]{0.49\columnwidth}
        \includegraphics[width=\columnwidth]{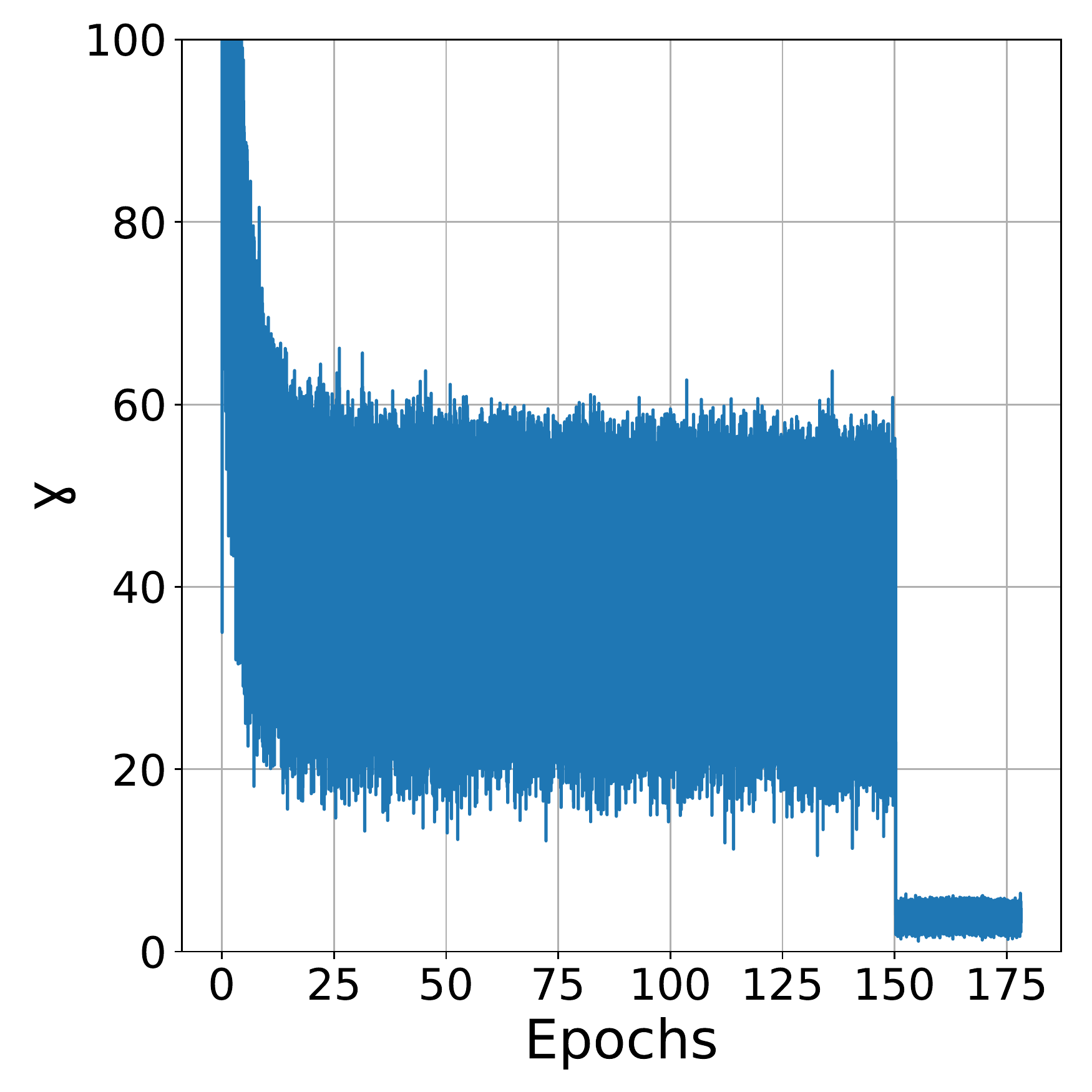}
        \centering
        \caption{QRNN --- PTB}        
        \label{fig:tiger}
    \end{subfigure}
    \caption{Evolution of the estimated SGD learning rate ($\gamma_k$) on two representative tasks. \label{fig:evolr}}
\end{figure}

The experiments comparing SGD, Adam and \name on the CIFAR and Tiny-ImageNet data sets are presented in Figures \ref{fig:cifars} and \ref{fig:tiny}, respectively. The experiments comparing the optimizers on the language modeling tasks are presented in Figure~\ref{fig:awdlstmlm}. In Table \ref{table:metaanalysis}, we summarize the meta-data concerning our experiments including the learning rates that achieved the best performance, and, in the case of \name, the number of epochs before the switch occurred and the learning rate ($\Lambda$) for SGD after the switch.  Finally, in Figure~\ref{fig:evolr}, we depict the evolution of the estimated SGD learning rate ($\gamma_k$) as the algorithm progresses on two representative tasks.

With respect to the image classification data sets, it is evident that, across different architectures, on all three data sets, Adam fails to find solutions that generalize well despite making good initial progress. This is in agreement with the findings of \cite{2017arXiv170508292W}. As can be seen from Table \ref{table:metaanalysis}, the switch from Adam to SGD happens within the first $20$ epochs for most CIFAR data sets and at epoch $49$ for Tiny-ImageNet. Curiously, in the case of the Tiny-ImageNet problem, the switch from Adam to SGD leads to significant but temporary degradation in performance. Despite the testing accuracy dropping from $80\%$ to $52\%$ immediately after the switch, the model recovers and achieves a better peak testing accuracy compared to Adam. We observed similar outcomes for several other architectures on this data set. 

In the language modeling tasks, Adam outperforms SGD not only in final generalization performance but also in the number of epochs necessary to attain that performance. This is not entirely surprising given that \citet{merityRegOpt} required iterate averaging for SGD to achieve state-of-the-art performance despite gradient clipping or learning rate decay rules. In this case, \name switches over to SGD, albeit later in the training process, but achieves comparable generalization performance to Adam as measured by the lowest validation perplexity achieved in the experiment. Again, as in the case of the Tiny-ImageNet experiment (Figure \ref{fig:tiny}), the switch may cause a temporary degradation in performance from which the model is able to recover.

%TODO: Which have momentum, which don't
%TODO: Which have momentum, which don't

These experiments suggest that it is indeed possible to combine the best of both worlds for these tasks: in all the tasks described, \name performs almost as well as the best algorithm amongst SGD and Adam, and in several cases achieves a good initial decrease in the error metric.

Figure~\ref{fig:evolr} shows that the estimated learning rate for SGD ($\gamma_k$) is noisy but convergent (in mean), and that it converges to a value of similar scale as the value obtained by tuning the SGD optimizer (see Table~\ref{table:metaanalysis}). We emphasize that other than the learning rate, no other hyperparameters were tuned between the experiments.

\begin{table*}
%\small
\center
\begin{tabular}{l|l|ccc|cr}
\toprule
Model & Data Set & SGDM & Adam & \name & $\Lambda$ & Switchover Point (epochs) \\
\midrule
ResNet-32 & CIFAR-10 & 0.1 & 0.001 & 0.001 & 0.52 & 1.37 \\
DenseNet & CIFAR-10 & 0.1 & 0.001 & 0.001 & 0.79 & 11.54 \\
PyramidNet & CIFAR-10 & 0.1 & 0.001 & 0.0007 & 0.85 & 4.94 \\
SENet & CIFAR-10 & 0.1 & 0.001 & 0.001 & 0.54 & 24.19 \\
ResNet-32 & CIFAR-100 & 0.3 & 0.002 & 0.002 & 1.22 & 10.42 \\
DenseNet & CIFAR-100 & 0.1 & 0.001 & 0.001 & 0.51 &  11.81 \\
PyramidNet & CIFAR-100 & 0.1 & 0.001 & 0.001 & 0.76 & 18.54 \\
SENet & CIFAR-100 & 0.1 & 0.001 & 0.001 & 1.39 & 2.04 \\
LSTM & PTB & 55$^\dagger$ & 0.003 & 0.003 & 7.52 & 186.03 \\ 
QRNN & PTB & 35$^\dagger$ & 0.002 & 0.002 & 4.61 &  184.14 \\ 
LSTM & WT-2 & 60$^\dagger$ & 0.003 & 0.003 & 1.11 & 259.47 \\ 
QRNN & WT-2 & 60$^\dagger$ & 0.003 & 0.004 & 14.46 &  295.71 \\ 
ResNet-18 & Tiny-ImageNet & 0.2 & 0.001 & 0.0007 & 1.71 & 48.91 \\
\end{tabular}
\caption{Summarizing the optimal hyperparameters for SGD(M), Adam and \name for all experiments and, in the case of \name, the value of the estimated learning rate for SGD after the switch and the switchover point in epochs. $\dagger$ denotes that no momentum was employed for SGDM.}
\label{table:metaanalysis}
\end{table*}

% \begin{table*}
% %\small
% \center
% \begin{tabular}{c|c|cc|cc|cc}
%  & & \multicolumn{2}{c|}{$\beta_2=0.99$} & \multicolumn{2}{c|}{$\beta_2=0.999$} & \multicolumn{2}{c}{$\beta_2=0.9999$} \\
%  Model & & Adam & \name &  Adam & \name & Adam & \name \\
% \toprule
% DenseNet on CIFAR-10 & $\beta_1=0$ \\
% & $\beta_1=0.9$ \\
% \midrule
% DenseNet on CIFAR-100 & $\beta_1=0$ \\
% & $\beta_1=0.9$ \\
% \midrule
% LSTM on PTB & $\beta_1=0$ \\
% & $\beta_1=0.9$ \\
% LSTM on WT-2 & $\beta_1=0$ \\
% & $\beta_1=0.9$ \\
% \end{tabular}
% \caption{Sensitivity to $\beta_1$ and $\beta_2$ for various models with learning rate set to the default value of $10^{-3}$.}
% \label{table:metaanalysis}
% \end{table*}

\section{Discussion and Conclusion}
\label{sec:discon}
% \begin{itemize}
% \item takes an orthogonal approach, leverages existing optimizers. might be interesting to club
% \item nothing binding it to Adam, should work well even with RMSprop or Adagrad
% \item back and forth adam and sgd
% \item smooth transition
% \item cyclic scheduling
% \item other strategies such as AMS or L2 norm fixing
% \item usual abstract stealing
% \end{itemize}
\citet{2017arXiv170508292W} pointed to the insufficiency of adaptive methods, such as Adam, Adagrad and RMSProp, at generalizing in a fashion comparable to that of SGD. In the case of a convex quadratic function, the authors demonstrate that adaptive methods provably converge to a point with orders-of-magnitude worse generalization performance than SGD. The authors attribute this generalization gap to the scaling of the per-variable learning rates definitive of adaptive methods as we explain below.  

Nevertheless, adaptive methods are important given their rapid initial progress, relative insensitivity to hyperparameters, and ability to deal with ill-scaled problems. Several recent papers have attempted to explain and improve adaptive methods \cite{2017arXiv171105101L,anonymous2018on,zhang2017normalized}. However, given that they retain the adaptivity and non-uniform gradient scaling, they too are expected to suffer from similar generalization issues as Adam. Motivated by this observation, we investigate the question of using a hybrid training strategy that starts with an adaptive method and switches to SGD. By design, both the switchover point and the learning rate for SGD after the switch, are determined as a part of the algorithm and as such require no added tuning effort. We demonstrate the efficacy of this approach on several standard benchmarks, including a host of architectures, on the PennTree Bank, WikiText-2, Tiny-ImageNet, CIFAR-10 and CIFAR-100 data sets. In summary, our results show that the proposed strategy leads to results comparable to SGD while retaining the beneficial properties of Adam such as hyperparameter insensitivity and rapid initial progress.  

The success of our strategy motivates a deeper exploration into the interplay between the dynamics of the optimizer and the generalization performance. Recent theoretical work analyzing generalization for deep learning suggests coupling generalization arguments with the training process \cite{soudry2017implicit,hardt2015train,zhang2016understanding,2017arXiv170508292W}. The optimizers choose different trajectories in the parameter space and are attracted to different basins of attractions, with vastly different generalization performance. Even for a simple least-squares problem: $\min_w \|Xw -y \|_2^2$ with $w_0=0$, SGD recovers the minimum-norm solution, with its associated margin benefits, whereas adaptive methods do not. The fundamental reason for this is that SGD ensures that the iterates remain in the column space of the $X$, and that only one optimum exists in that column space, viz. the minimum-norm solution. On the other hand, adaptive methods do not necessarily stay in the column space of $X$. Similar arguments can be constructed for logistic regression problems \cite{soudry2017implicit}, but an analogous treatment for deep networks is, to the best of our knowledge, an open question. We hypothesize that a successful implementation of a hybrid strategy, such as \name, suggests that in the case of deep networks, despite training for few epochs before switching to SGD, the model is able to navigate towards a basin with better generalization performance. However, further empirical and theoretical evidence is necessary to buttress this hypothesis, and is a topic of future research. 

While the focus of this work has been on Adam, the strategy proposed is generally applicable and can be analogously employed to other adaptive methods such as Adagrad and RMSProp. A viable research direction includes exploring the possibility of switching back-and-forth, as needed, from Adam to SGD. Indeed, in our preliminary experiments, we found that switching back from SGD to Adam at the end of a 300 epoch run for any of the experiments on the CIFAR-10 data set yielded slightly better performance. Along the same line, a future research direction includes a smoother transition from Adam to SGD as opposed to the hard switch proposed in this paper, which may cause short-term performance degradation. This can be achieved by using a convex combination of the SGD and Adam directions as in the case of \citet{akiba2017extremely}, and gradually increasing the weight for the SGD contribution by a criterion. Finally, we note that the strategy proposed in this paper does not preclude the use of those proposed in \citet{zhang2017normalized,2017arXiv171105101L,anonymous2018on}. We plan to investigate the performance of the algorithm obtained by mixing these strategies, such as monotonic increase guarantees of the second-order moment, cosine-annealing, $\ell_2$-norm correction, in the future.	  

%TODO: Make some commentary regd. not too late

\bibliography{example_paper}
\bibliographystyle{icml2017}

\end{document}